\documentclass{article}

\usepackage{arxiv}

\usepackage[utf8]{inputenc} 
\usepackage[T1]{fontenc}    
\usepackage{hyperref}       
\usepackage{url}            
\usepackage{booktabs}       
\usepackage{amsfonts}       
\usepackage{nicefrac}       
\usepackage{microtype}      
\usepackage{lipsum}
\usepackage{graphicx}
\usepackage{amsmath,amssymb}
\usepackage{url}
\usepackage{graphicx}
\usepackage{float} 
\usepackage{multirow}
\usepackage{color, colortbl}
\definecolor{b}{RGB}{50,153,204}
\definecolor{RawSienna}{RGB}{229, 108, 76}
\definecolor{Gray}{gray}{0.9}
\usepackage{booktabs}
\usepackage{float} 
\usepackage[super]{nth}
\newtheorem{definition}{Definition}
\graphicspath{ {./images/} }

\title{A study of the Multicriteria decision analysis based on the time-series features and a TOPSIS method proposal for a tensorial approach.}

\author{Betania S. C. Campello \\
  School of Electrical and Computer Engineering (FEEC)\\
  University of Campinas\\
  betania@decom.fee.unicamp.br\\
   \And
 Leonardo T. Duarte \\
  School of Applied Sciences (FCA)\\
  University of Campinas\\
  leonardo.duarte@fca.unicamp.br\\
  \And
 João M. T. Romano \\
  School of Electrical and Computer Engineering (FEEC)\\
  University of Campinas\\
  leonardo.duarte@fca.unicamp.br\\
}

\begin{document}
\maketitle
\begin{abstract}
A number of Multiple Criteria Decision Analysis (MCDA) methods have been developed to rank alternatives based on several decision criteria. Usually, MCDA methods deal with the criteria value at the time the decision is made without considering their evolution over time. However, it may be relevant to consider the criteria' time series since providing essential information for decision-making (e.g., an improvement of the criteria). To deal with this issue, we propose a new approach to rank the alternatives based on the criteria time-series features (tendency, variance, etc.).  In this novel approach, the data is structured in three dimensions, which require a more complex data structure, as the \textit{tensors}, instead of the classical matrix representation used in MCDA. Consequently, we propose an extension for the TOPSIS method to handle a tensor rather than a matrix. Computational results reveal that it is possible to rank the alternatives from a new perspective by considering meaningful decision-making information. 

\end{abstract}

\section{Introduction}
\label{sec:introduction}

Many cases require the decision-maker to rank alternatives according to multiple decision criteria. When this decision requires dealing with a significant amount of data, methods of multiple criteria decision analysis (MCDA) arise as an interesting tool~\cite{roy1985methodologie, e2012readings, mardani2015multiple}. These methods are often used to rank a set of alternatives $A = \{a_1, \ldots, a_m\}$ based on a set of criteria $C = \{c_1, \ldots, c_n\}$~\cite{keeney1993decisions}. MCDA methods are applied in several fields, including the public sector~\cite{dotoli2020multi}, sustainable development~\cite{frini2019mupom}, economics and finance~\cite{masri2018financial}, medicine and health care~\cite{belacel2000multicriteria, oliveira2019multi}, energy storage systems~\cite{baumann2019review}, and many others~\cite{zopounidis2002multicriteria}.
 
Most of MCDA techniques consider as input data the \textit{decision matrix} \textbf{P} $\in \mathbb{R}^{m \times n}$. The matrix rows represent alternatives to be ranked and the columns represent criteria.  The performance of the alternatives is measure in terms of the values assumed concerning the criteria. Each element $p_{ij}$ of matrix \textbf{P} corresponds to the evaluation of alternative $i$ in criterion $j$. A core issue in MCDA, known as \textit{matrix aggregation}, consists of applying a technique that transforms the decision matrix into a scoring vector \textbf{g}; i.e., each row $i$ of the matrix \textbf{P} (the alternatives) is mapped into a score $g_i$, used to rank the alternatives. 

MCDA approaches usually consider a static value ($p_{ij}$) to evaluate  alternative $i$ concerning criterion $j$, without dealing with the criteria evolution over time. The $p_{ij}$ is often the value at the time that the decision is taken (\textit{current data}). Although this approach is widely used, many decision-making problems require analyzing the time-series features beyond only the current data. For instance, it can be relevant to consider the variance, the tendency, the seasonality, among other time-series features of the criteria.

\begin{figure}[h]
	\centering
	\includegraphics[width=10cm]{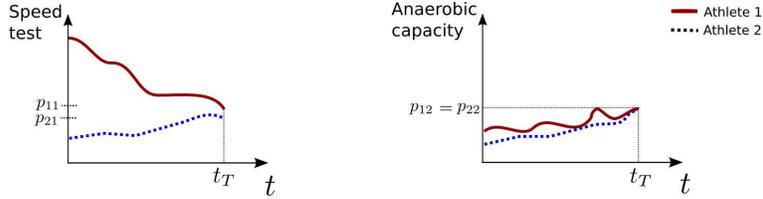}
	\caption{Analysis of the features of the criteria.}
	\label{fig:graf}
\end{figure}

To illustrate the interest behind the analysis of the time-series information in MCDA, let us suppose that we aim to rank two athletes based on their speed test and anaerobic capacity, as shown in Figure~\ref{fig:graf}. If we ponder only the current data ($t_T$), Athlete 1 is chosen due to its speed test superiority. Likewise, Athlete 1 is chosen if we consider all the values one by one over the time-series. However, it is interesting to note that Athlete 2 is improving in both criteria, whereas Athlete 1 is worsening in the speed test and presents less  regularly  in anaerobic capacity. Hence, the decision can be made from a new perspective in which athletes' improvement and regularity are considered. In this case, Athlete 2 should be chosen due to the high slope coefficient and low variance in both criteria. This example suggests that different rankings (i.e., different solutions) can be achieved by taking into account the time-series features, as we shall discuss later.

Few studies in MCDA regard the time-series of the criteria. For example,~\cite{Mingl2011} investigated the decision information at different periods applied in emergency management with real-world time-series. The authors proposed the Ordered Weighted Averaging method for aggregating the time-series.~\cite{frini2015topsis} and~\cite{frini2019mupom} considered the criteria time-series in a sustainable development context. The authors applied an aggregating method called Multi-criteria multi-Period Outranking Method. Other authors, such as~\cite{banamar2018extension} and \cite{campello2020adaptive}, also analyzed time-series in the MCDA approach. These studies differ from our proposal since they apply a method to aggregate the time-series; instead, we aim to explore many time-series' features before the aggregation.     
 
\begin{figure}[h]
	\centering
	\includegraphics[width=12cm]{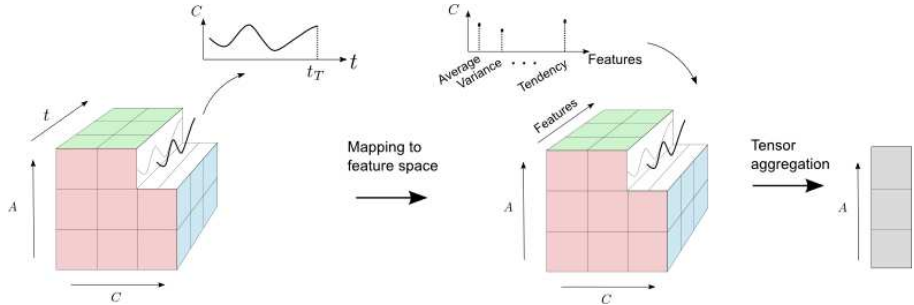}
	\caption{Tensorial representation, space mapping, and tensor aggregation.}
	\label{fig:agregacao_tensor}
\end{figure}

Therefore, in this study we analyze the MCDA problem dynamically by representing a given criterion as a time-series (signal). This approach leads us to structure the data involved in the decision as a \textit{tensor}~\cite{sidiropoulos2017tensor, da2018tensor} (a function of three or more indexes), as shown in Figure~\ref{fig:agregacao_tensor}. Our first contribution consists of obtaining features of the signals that may be relevant for the decision. In other words, we mapped the time-space into a feature-space by taking measures that describe the time-series evolution. From this mapping, the third tensor dimension  becomes the time-series feature. Finally, we apply a method to aggregate the tensor for ranking the alternatives.
 
Since most MCDA methods deal with a matricial structure, we propose an MCDA method extension to provide a tensor aggregation. Thus, our second contribution is to extend the procedure called Technique for Order of Preference by Similarity to Ideal Solution (TOPSIS)~\cite{hwang1981methods} for aggregate the tensor of features. The classical TOPSIS method considers that the best alternative should have the shortest distance of the \textit{ideal positive point} and the greatest distance of the \textit{ideal negative point}~\cite{behzadian2012state}. The ideal positive point is the best value the criteria assume for all alternatives, and the ideal negative point, the worse one. This method is one of the most popular MCDA techniques, as it presents a good performance and is suitable for different application areas~\cite{behzadian2012state}.  


Many extensions and applications of the TOPSIS have been proposed, such as found in~\cite{jahanshahloo2006algorithmic},~\cite{chen2015inclusion},~\cite{frini2015topsis},~\cite{dash2019integrated},~\cite{palczewski2019fuzzy}, and~\cite{shukla2017applications}. An aspect of the TOPSIS method and its extensions is that it incorporates relative weights that model each criterion's importance~\cite{olson2004comparison}.  Our proposal introduces relative weights for modeling the features' importance since they can have different relevance depending on the decision's objective. For instance, in the example of the athletes' ranking, a positive trend in athlete performance may be more relevant than their performance variance. 
  

Finally, a sensitivity analysis is given in the computational tests to analyze our proposal's performance further. We study changes in the alternatives' ranking when the feature weights are modified. For this purpose, we use the stochastic multicriteria acceptability analysis (SMAA) method technique~\cite{lahdelma1998smaa}. The SMAA is widely used for support decision-makers to explore the weight space. It can be used in contexts where the weights are uncertain, or the decision-makers do not express their weights preferences. The central idea is to obtain the ranking for each possible value of the parameter, e.g., calculate the ranking for each weight vector value. The method provides the probability of each alternative being in each position for all possible parameter values. Thus, from the SMAA output, it is possible to observe if the specific ranking is more likely, or if there is more probability of a given alternative to be the preferred one, among other analyses. 

The paper is organized as follows. Subsection~\ref{sec:tensors} summarizes the tensor notation used in this paper. Section~\ref{sec:flex} discusses the motivation of considering the feature-space. Section~\ref{sec:methodology} describes the methodology of this study, and it is divided into two subsections: Subsection~\ref{sec:TOPSIS} describes the TOPSIS method extension and Subsection~\ref{sec:smaa}, the SMAA method. In Section~\ref{sec:resultados}, the results and their discussion are given. Finally, Section~\ref{sec:conclusoes} concludes this study.

\subsection{Mathematical notation}
\label{sec:tensors}

\begin{figure}[h]
	\centering
	\includegraphics[width=12cm]{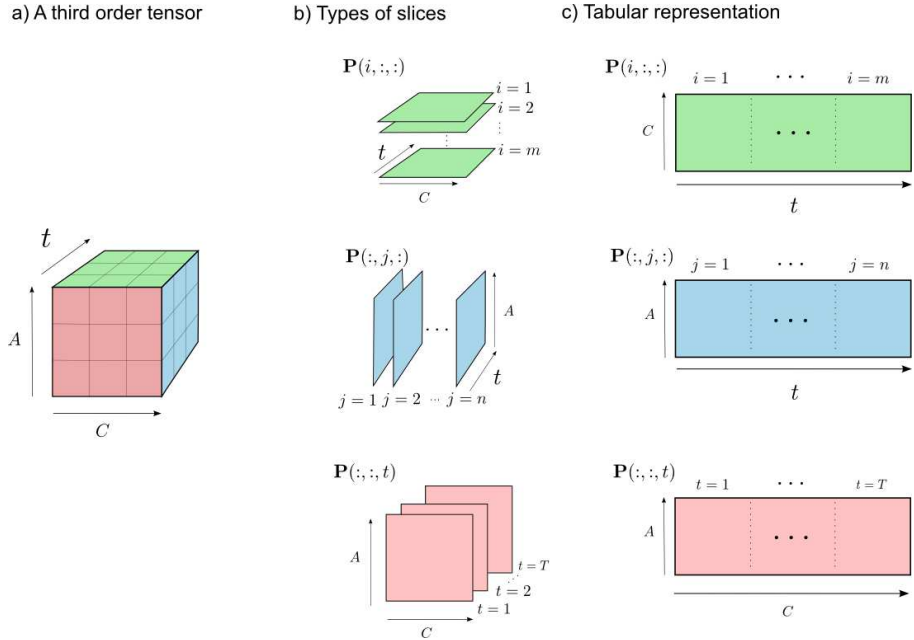}
	\caption{Slices-types for a third order tensor and its tabular representation.}
	\label{fig:cortes_tensor}
\end{figure}

We briefly present the tensor notations used in this paper. For further details, we refer the interested reader to~\cite{cichocki2015tensor, da2018tensor, kanatsoulis2019regular}. A real-valued tensor is denoted by $\mathcal{P} \in \mathbb{R}^{K_1 \times K_2 \times \cdots \times K_o}$, where $o$ is its order (number of dimensions). The elements of the tensor are represented by $p_{k_1 \ldots k_o}$. Subtensors are obtained by fixing a subset of tensor indices. The matrix subtensors, or \textit{slices}, are defined by fixing all but two indices.  A third order tensor $\mathcal{P} \in \mathbb{R}^{m \times n \times T}$ has three slices: vertical $ \textbf{P}(i, :, :) $, horizontal $ \textbf{P}(:, j, :) $ and frontal $ \textbf{P}(:, :, t)$, as shown in Figure~\ref{fig:cortes_tensor}.

\section{The motivation of considering feature-space}
\label{sec:flex} 
  
As discussed in Section~\ref{sec:introduction}, most MCDA studies consider the criteria value as the current data, and few studies deal with the criteria' time-series~\cite{frini2017making}. That is, both approaches ranking the alternatives taking into account the criteria value in the time domain. However, the this domain is not always the most favorable to analyze relevant aspects of the decision. It can be more appropriate to rank the alternatives considering the criteria in a feature domain in some decisions. This domain transformation may lead to different solutions.

The latter statement becomes clearer in the example to rank athletes used in Section~\ref{sec:introduction}. We observed that Athlete 1 should be preferred, whether it is considered the criteria' current data. Also, Athlete 1 is chosen considering all values one by one over the time-series. Note that both approaches are in the time domain. Instead, by considering the improvement in the athletes' performance (the slope coefficient of the criteria' time-series), Athlete 2 should be chosen. That is, by mapping the time-space into a feature-space, a different solution is achieved. The solution with the feature-space approach may be interesting, for instance, to hire an athlete. Since Athlete 2 performance is improving, it shall be better than Athlete 1 performance in the medium term. 

Before further discuss the change of ranking when a feature domain is considered, let us define concepts relevant to MCDA~\cite{grabisch2009aggregation}:
  
\begin{definition}
	A function $f: \mathbb{R}^{n} \rightarrow \mathbb{R}$ is monotonically nondecreasing in each argument if, for any vector $\textbf{p}_1$, $\textbf{p}_2$ $\in \mathbb{R}^n$, $\textbf{p}_1 \geq \textbf{p}_2 \implies f\{\textbf{p}_1\} \geq f\{\textbf{p}_2\}$. 
	
\end{definition}

\begin{definition}\label{def:agg_fun}
	An aggregation function in $\mathbb{R}^{m \times n}$ is a function $f: \mathbb{R}^{m \times n} \rightarrow \mathbb{R}^n$, which a natural requirement is nondecreasing monotonicity in each argument.
\end{definition}

\begin{definition}
	The decision criteria can be either benefit (maximum) when the desired value is as high as possible, or cost (minimum) when the desired value is as low as possible.
\end{definition}

 
Following the example of ranking the athletes, suppose that the decision-maker should choose one among two alternatives ($m = 2$), according to two criteria ($n = 2$) of benefit, where each criterion is measured over two periods $T = 2$. Let us represent the decision data as a tensor $\mathcal{P} \in \mathbb{R}^{2 \times 2 \times 2}$, in which each alternative  $i$ is represented by vertical slices of $\mathcal{P}$,  $ \textbf{P}(i, :, :) \in \mathbb{R}^{2 \times 2} $:
\begin{equation}
\label{eq:matr_ex}
 \textbf{P}(i, :, :)=\begin{array}{cc} 
\left[\begin{array}{cc}
p_{i11} & p_{i12}   \\
p_{i21} & p_{i22} 
\end{array}
\right].
\end{array}
\end{equation} 

\begin{figure}[h]
	\centering
	\includegraphics[height=5cm,keepaspectratio]{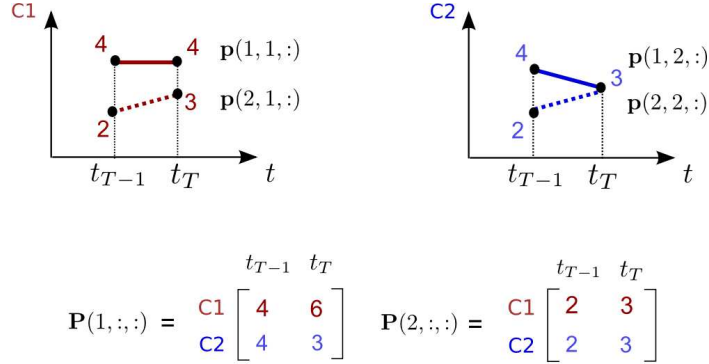}
	\caption{Data involved in the decision-making.}
	\label{fig:decision_data}
\end{figure}

\noindent Each vector $ \textbf{p}(i,j,:) $ represents the  time-series of alternative $i$ in criterion $j$. Figure~\ref{fig:decision_data} shows the data and graphics.  

By Figure~\ref{fig:decision_data} it is possible to see that the current data of Alternative 1 are greater than or equal to the current data of Alternative 2, i.e., $p_{1jT} \geq p_{2jT}$ $\forall j$. Also, all the values of Alternative 1 in the time-series are greater than or equal to the values of Alternative 2, $p_{ijt} \geq p_{ijt}$ $\forall j,  t$. According to \textbf{Definition 1} and \textbf{Definition 2} by applying any aggregation function $f\{.\}$ in $\textbf{P}(1, :, :)$ and $\textbf{P}(2, :, :)$, since $p_{ijt} \geq p_{ijt}$ $\forall j, t$, implies that  $f\{\textbf{P}(1, :, :)\} \geq f\{\textbf{P}(2, :, :)\}$. That means Alternative 1 is preferable to Alternative 2 independent of the MCDA aggregation method. Thus, Alternative 1 dominates Alternative 2 in the time domain. Suppose $f\{.\}$ an additive method: 

\begin{eqnarray}\label{eq:frobenius}
f\{\textbf{P}(i, :, :)\} = \sum_{j = 1}^{n} \left(\sum_{t = 1}^{T}w_{tj} p_{ijt} \right),  \  i = 1,\cdots, m, 
\end{eqnarray} 

\noindent where the elements $w_{tj}$ of a matrix \textbf{W}, models the relative importance of criterion $j$ in period $t$, and $w_{tj} \geq 0$ and  $\sum_{t=1}^{T} \sum_{j=1}^{n}   w_{tj} = 1$. Assuming:
\begin{equation} 
\label{eq:matr_pesos}
\textbf{W}=\begin{array}{cc}  
\left[\begin{array}{cc}
0.25 & 0.25   \\
0.25 & 0.25   
\end{array}
\right],
\end{array}
\end{equation} 

\noindent by Equation~(\ref{eq:frobenius}), $f\{\textbf{P}(1, :, :)\} = 4$ and $f\{\textbf{P}(2, :, :)\} = 2.5$. As expected, Alternative 1 is preferable to Alternative 2:
\begin{equation}\label{eq:ranking}
	\textbf{P}(1, :, :) \succ \textbf{P}(2, :, :),
\end{equation}

\noindent Same solution is obtained if we use only the current data.  Notice that it is not possible to find \textbf{W} able to change this ranking because $p_{1jt} \geq p_{2jt}$ $\forall j, t$, which implies that $\sum_{j=1}^{n} \sum_{t=1}^{T}p_{1jt} \geq \sum_{j=1}^{n} \sum_{t=1}^{T} p_{2jt}$; finally, since the weights are non-negative $w_{tj} \geq 0$, by multiplying non-negative numbers in both sides of the inequality, it is not possible to change the ranking: $\sum_{j=1}^{n} \sum_{t=1}^{T}w_{tj} p_{1jt} \geq \sum_{j=1}^{n} \sum_{t=1}^{T} w_{tj}p_{2jt}$.    

Suppose now that it is relevant in the decision to consider the tendency of the criteria. Therefore, a change of domain is made by mapping the time-space into a feature-space. This mapping is represented by $G\{.\}$: $G\{\textbf{P}(1, :, :)\}$ and $G\{\textbf{P}(2, :, :)\}$. Let assume $G(.)$ as the slope coefficient. We apply the mapping $G\{.\}$, and then apply the additive method $f\{.\}$ defined in Equation~\ref{eq:frobenius}, obtaining $f\{G\{\textbf{P}(1, :, :)\}\} = -0.25$ and $f\{G\{\textbf{P}(2, :, :)\}\}= 0.5$. Hence, Alternative 2 becomes preferable to  Alternative 1:
\begin{equation}\label{eq:ranking2}
	\textbf{P}(2, :, :) \succ \textbf{P}(1, :, :).
\end{equation}
 
The preference in~(\ref{eq:ranking2}) is different from~(\ref{eq:ranking}). In the time domain, Alternative 1 dominates Alternative 2. In a feature space, Alternative 2 becomes the preferred one. The rankings are different, but both are suitable to support the decision-maker. Thus, the example shows that considering only the time domain can disregard a solution that can be interesting in the feature domain. In other words, to ignore the features leads to the loss of crucial information and the solution is not necessarily satisfactory. Therefore, our proposal considers more elements besides the current data, such as tendency, mean, variance, etc. The next section presents the methodology proposed to achieve this objective.
 
\section{Methodology}
\label{sec:methodology}

\subsection{Extended TOPSIS method for aggregate the tensor}
\label{sec:TOPSIS}

Figure~\ref{fig:agregacao_tensor} of Section~\ref{sec:introduction} shows the tensorial representation used to structure the data in the feature-space. This novel data representation requires to adapt the MCDA methods in other to aggregate the tensor and ranking the alternatives. This section introduces the proposed extension of the TOPSIS method~\cite{hwang1981methods} for a tensorial approach. The classical TOPSIS consists of measuring the distance of the ideal positive and negative points. The ideal positive point of a specific criterion is the best value of this criterion in all alternatives, and the ideal negative point is the worst one. The best value is the highest value if the criterion is of benefit and the lowest if it is of cost. Therefore, the scoring value of each alternative is determined according to the criteria values' distance to the positive and negative ideal points. The data input in this method is the decision matrix and the vector of weights \textbf{w}.
 
In the TOPSIS extension we propose, the algorithm input data is a third-order tensor $\mathcal{P}  \in \mathbb{R}^{m \times n \times T}$ , where $m$ is the number of alternatives, $n$ is the number of criteria, and $T$ the number of samples in the time-series. In this extension, we first map the time-space into a feature-space $\mathcal{P}  \in \mathbb{R}^{m \times n \times T} \Rightarrow \mathcal{S}  \in \mathbb{R}^{m \times n \times h}$, where $h$ is the number of features. The criteria weights are  $w_j, \  j = 1,\ldots, n$, and $w_j \geq 0$, $\sum_{j=1}^{n} w_j = 1$. We also introduce weights for the features, represented by $\alpha_k, \ k = 1,\ldots, h$, in which  $ \alpha_k \geq 0$, $\sum_{k=1}^{h} \alpha_k = 1$. 

Finally, we verify if the feature $k$ is of benefit or cost to determine the ideal positive and negative point. This step is necessary since some features are  desirable to be as high (or low) as possible, independent of the criteria. For instance, if a low variance is important for the decision, independent of the criteria, the ideal positive point will be the lowest. However, the tendency is neither benefit nor cost, since it depends on whether the criterion is of benefit or cost; if the criterion is of benefit, we desire an increasing tendency. If the criterion is of cost, we desire to decrease tendency.

In the sequence, we detail the steps of the proposed extension of the TOPSIS.

\begin{enumerate}
	\item Map the time-space into a feature-space. 
	\begin{eqnarray}
	\mathcal{P}  \in \mathbb{R}^{m \times n \times T} \Rightarrow \mathcal{S}  \in \mathbb{R}^{m \times n \times h},
	\end{eqnarray}
	where the $s_{ijk}$ are the elements of $\mathcal{S}$ and $h$ the number of features. The chosen features should be those that are relevant to the purpose of the decision.
	
	\item Normalize the elements of the tensor $\mathcal{S}$, which results in the tensor represented by $\mathcal{N}$.
	\begin{eqnarray}
	n_{ijk}=  \frac{s_{ijk}}{\sqrt{\sum_{j=1}^{n}(s_{ijk})^2}}, \ \ i = 1,\ldots, m, \ j = 1,\ldots, n, \ k = 1,\ldots, h.
	\end{eqnarray}
	This step is necessary since the order of magnitude and the unit of measurement of the data may influence the results.
	
	\item Weight the tensor $\mathcal{N}$, from where we obtain the tensor represented by $\mathcal{V}$.
	\begin{eqnarray}
	v_{ijk}= \alpha_k n_{ijk}w_{j}, \ \ i = 1,\ldots, m, \ j = 1,\ldots, n, \ k = 1,\ldots, h.
	\end{eqnarray}

	\item Determine the positive and negative ideal points represented by $\mathcal{A^+}$ e $\mathcal{A^-}$ $\in \mathbb{R}^{1 \times n \times h}$. The ideal positive and negative points depend on if the feature  $k$ is in the set of benefit $I$ or cost $J$ (by Equation~(\ref{eq:beneforcost}), where `$\vee$' represents the logical operator \textit{or}); or if the feature $k$ is neither benefit nor cost (by Equation~(\ref{eq:benefecost}), where `$\wedge$' represents the logical operator \textit{and}). If it is of benefit, we identify the highest value that the alternatives assume for each feature and each criterion. If it is of cost, we determine the lowest value that the alternatives assume for for each feature and each criterion. If the feature is neither benefit nor cost, we identify if the criterion $j$ is of benefit $I$ or cost $J$. Thus, for each $k = 1,\ldots, h$:
	\begin{flalign} 
	k \in I  \vee k \in J \rightarrow  \label{eq:beneforcost} \\
	A_k^+ = \{v_{1k}^+, \ldots, v_{nk}^+ \} = \bigg\{\bigg(\underset{j}{\max} \  v_{ijk} \mid k \in I\bigg), \bigg(\underset{j}{\min} \  v_{ijk} \mid k \in J\bigg)\bigg\} \\ 
	A_k^- = \{v_{1k}^-, \ldots, v_{nk}^- \} = \bigg\{\bigg(\underset{j}{\min} \  v_{ijk} \mid k \in I\bigg), \bigg(\underset{j}{\max} \  v_{ijk} \mid k \in J\bigg)\bigg\} \\ 
	k \not\in I \wedge \ k \not\in J \rightarrow \label{eq:benefecost} \\
	A_k^+ = \{v_{1k}^+, \ldots, v_{nk}^+ \} = \bigg\{\bigg(\underset{j}{\max} \  v_{ijk} \mid j \in I\bigg), \bigg(\underset{j}{\min} \  v_{ijk} \mid j \in J\bigg)\bigg\} \   \\
	A_k^- = \{v_{1k}^-, \ldots, v_{nk}^- \} = \bigg\{\bigg(\underset{j}{\min} \  v_{jik} \mid j \in I\bigg), \bigg(\underset{j}{\max} \  v_{jik} \mid j \in J\bigg)\bigg\};
	\end{flalign}
	 
	\item  Compute the $\mathit{n}$-dimensional Euclidean distance of each alternative for the ideal positive and negative points.
	\begin{eqnarray}
	d_i^+ = \Bigg\{\sum_{k=1}^{h}\sum_{j=1}^{n}(v_{ijk}-v^+_{jk})^2\Bigg\}^{\frac{1}{2}}, \ i = 1,\ldots, m\\
	d_i^- = \Bigg\{\sum_{k=1}^{h}\sum_{j=1}^{n}(v_{ijk}-v^-_{jk})^2\Bigg\}^{\frac{1}{2}}, \ i = 1,\ldots, m;
	\end{eqnarray} 
	
	\item Compute the relative proximity of the alternatives to the positive optimal point, where the vector $\textbf{g} = [g_1, \ldots, g_m] $ is obtained.
	\begin{eqnarray}
	g_i = \frac{d_i^-}{d_i^+ + d_i^-}, \ i = 1,\ldots, m.
	\end{eqnarray}
	By sorting in descending order the values $g_i$ in \textbf{g}, we obtain the ranking of the alternatives. Each $g_i$ assumes values between zero and one. When $g_i$ tends to 1, the alternative tends to be closer to the ideal positive point and far from the negative ideal.  
\end{enumerate}

\subsection{Stochastic multicriteria acceptability analysis (SMAA)}
\label{sec:smaa}
 
In Section~\ref{sec:flex}, we showed a change of ranking when the criteria value is in the time domain or by considering the slope coefficient. In the computational test, we deepen this analysis adding  more features besides the slope coefficient. Because in the TOPSIS method the weights of the features are considered,  we provide a sensitivity analysis of these weights in the ranking of the alternatives; i.e., we use the SMAA method to verify changes in the ranking by varying  the weights of the features.

The SMAA numerical  calculation is quite complicated, but it can be replaced by Monte Carlo simulation, which generates good approximations~\cite{tervonen2007implementing}. In this study, the SMAA input is the feature-space tensor, $\mathcal{P}  \in \mathbb{R}^{m \times n \times T}$. We set the value of the criteria weights $w_j, \  j = 1,\ldots, n$.  The feature weights are represented by a random variable, $\alpha_k \sim U[a, b]$, for $k = 1,\ldots, h$,  where $U$ is a continuous uniform distribution, $0 \leq a < b \leq 1$, and $\sum_{k=1}^{h} \alpha_k = 1$. The called \textit{percentage matrix of rankings}, $ \textbf{M} = \textbf{0}_{m \times m} $ computes the percentage the alternatives occupied a given position. The elements of this matrix are represented by $m_{i\theta}$. The simulation consists of repeat the steps below $L$ times, where $L$ is a large number (in the order of a thousand): 
\begin{enumerate}
	\item Sample the random variables $\alpha_k$, for $k = 1,\ldots, h$, in other to compose the deterministic vector $\boldsymbol{\alpha}$;
	
	\item Then, apply the extension of the TOPSIS method presented in Section~\ref{sec:TOPSIS} with the inputs $\mathcal{P}$, \textbf{w}, and the weights $\boldsymbol{\alpha}$ obtained in Step 1. The output of the TOPSIS is a ranking \textbf{g};
	
	\item Compute the position of each alternative in matrix \textbf{M}. For this, do $m_{i\theta} = m_{i\theta} + 1$ if alternative $i$ is in position $\theta$ in the ranking \textbf{g}.	 
\end{enumerate}
\noindent At the end of the simulation, we compute the percentage: \textbf{M} = \textbf{M}/$L$ * 100. 

Each $m_{i\theta}$ represents the percentage of times that alternative $ i $ was at position $\theta$. If the alternative $i$ has the highest percentage in the $\theta$-th position, for all $i$ and $\theta$, of \textbf{M}, then it is possible to find the most likely ranking.

\section{Experiments on actual data and discussion}
\label{sec:resultados}
 
In this section, the Human Development Index (HDI) is calculated using the feature-space proposal. The HDI can be computed by aggregating three criteria: life expectancy at birth ($c_1$), education ($c_2$), and gross national income per capita ($c_3$). Usually, the criteria values are that they assume at the year the index is calculated. In~\cite{banamar2018extension} the authors proposed to calculated the HDI for ranking ten emerging economies: Brazil (BR), China (CN), India (IN), Indonesia (ID), Malaysia (MY), Mexico (MX), Philippines (PH), Russia (RU), South Africa (ZA), Turkey (TR); according to the criteria $c_1$, $c_2$, and $c_3$. They considering the evolution of the criteria over the years (the criteria time-series). To test our proposal, we consider the same data, which is shown in Table~\ref{tab:dados} in horizontal representation $\textbf{P}(:,j,:)$. The weights the criteria assumes are equal for all of them, i.e., $w_j = 0.333 $. 
 
\begin{table}[htbp]
	\centering
	\caption{Evaluations Table for 10 emerging countries~\cite{banamar2018extension}.}
	\scalebox{0.5}{\begin{tabular}{|c|r|r|r|r|r|r|r|r|r|r|r|r|r|r|r|r|r|r|}
			\hline
			Weights \textbf{w} & \multicolumn{ 6}{c|}{0.333} & \multicolumn{ 6}{c|}{0.333} & \multicolumn{ 6}{c|}{0.333} \\ \hline
			Max/Min & \multicolumn{ 6}{c|}{Max} & \multicolumn{ 6}{c|}{Max} & \multicolumn{ 6}{c|}{Max} \\ \hline
			Criteria & \multicolumn{ 6}{c|}{Life expectancy at birth ($c_1$)} & \multicolumn{ 6}{c|}{Education ($c_2$)} & \multicolumn{ 6}{c|}{Gross national income per capita ($c_3$)} \\ \hline
			\rowcolor{Gray}
			Years  & 1990 & 1995 & 2000 & 2005 & 2010 & 2015 & 1990 & 1995 & 2000 & 2005 & 2010 & 2015 & 1990 & 1995 & 2000 & 2005 & 2010 & 2015 \\ \hline
			BR & 65.3 & 67.6 & 70.1 & 71.9 & 73.3 & 74.8 & 8.00 & 8.95 & 9.95 & 10.15 & 11.05 & 11.60 & 10065 & 10959 & 11161 & 12032 & 14420 & 15062 \\ \hline
			CN & 69.0 & 69.9 & 71.7 & 73.7 & 75.0 & 76.0 & 6.80 & 7.25 & 7.85 & 8.75 & 9.85 & 10.30 & 1520 & 2508 & 3632 & 5632 & 9387 & 13347 \\ \hline
			IN & 57.9 & 60.4 & 62.6 & 64.5 & 66.5 & 68.4 & 5.35 & 5.90 & 6.45 & 7.35 & 8.25 & 8.55 & 1754 & 2046 & 2522 & 3239 & 4499 & 5814 \\ \hline
			ID & 63.3 & 65.0 & 66.3 & 67.2 & 68.1 & 69.1 & 6.75 & 7.20 & 8.70 & 9.30 & 9.95 & 10.30 & 4337 & 5930 & 5308 & 6547 & 8267 & 10130 \\ \hline
			MY & 70.7 & 71.8 & 72.8 & 73.6 & 74.1 & 74.8 & 8.10 & 8.90 & 10.25 & 10.15 & 11.35 & 11.35 & 9772 & 13439 & 14500 & 17157 & 19725 & 23712 \\ \hline
			MX & 70.8 & 72.8 & 74.4 & 75.3 & 76.1 & 77.0 & 8.05 & 8.55 & 9.15 & 9.85 & 10.50 & 10.80 & 12074 & 12028 & 14388 & 14693 & 15395 & 16249 \\ \hline
			PH & 65.3 & 66.1 & 66.7 & 67.2 & 67.7 & 68.3 & 8.70 & 8.95 & 9.50 & 9.75 & 9.75 & 10.20 & 3962 & 4111 & 4994 & 6058 & 7478 & 8232 \\ \hline
			RU & 68.0 & 66.0 & 65.1 & 65.8 & 68.6 & 70.3 & 10.95 & 10.85 & 11.85 & 12.60 & 13.10 & 13.35 & 19461 & 12011 & 12933 & 17797 & 21075 & 22094 \\ \hline
			ZA & 62.1 & 61.4 & 55.9 & 51.6 & 54.5 & 57.9 & 8.95 & 10.65 & 11.00 & 11.15 & 11.55 & 11.75 & 9987 & 9566 & 9719 & 10935 & 11833 & 12110 \\ \hline
			TR & 64.3 & 67.0 & 70.0 & 72.5 & 74.2 & 75.6 & 6.70 & 7.20 & 8.30 & 8.95 & 10.55 & 11.05 & 10494 & 11317 & 12807 & 14987 & 16506 & 18976 \\ \hline
		\end{tabular}}
		\label{tab:dados} 
	\end{table} 
	
The feature-space is composed of four features ($h = 4$), current data (2015 data), average, coefficient of variation (CV), and slope coefficient (SC).  Table~\ref{tab:atributos} shows the tensor $ \mathcal{S} \in \mathbb{R}^{10 \times 3 \times 4}$, in the  frontal representation $ \textbf{S}(:,:,k) $, obtained by computing these features. The CV is a feature of cost (CV $\in J$), and the features  SC, current data and average do not belong to either set of benefit or cost. 

The results are presented using five strategies. Each of the first four strategies deals with only one feature. From these four strategies, we can better analyze the countries position in the ranking given a specific feature. In the fifth strategy, all the features are considered, which is effectively our proposal. In this latter strategy, we use the SMAA method to support the choice of the weights of the features and to provide a sensitivity analysis concerning these features. The five strategies were implemented as follow:
\begin{description}  
	\item[Strategy 1:] In this strategy $\boldsymbol{\alpha}^{S1}$ = [1, 0, 0, 0], which means only current data is taking into account, equivalent to consider only the slice $ \textbf{S}(:,:,1)$. Notice that this strategy is as it has been used in the literature.
	
	\item[Strategy 2, 3 and 4:] In these three strategies, the $ \boldsymbol{\alpha}$ values are: $\boldsymbol{\alpha}^{S2}$, $\boldsymbol{\alpha}^{S3}$, $\boldsymbol{\alpha}^{S4}$ = [0, 1, 0, 0], [0, 0, 1, 0], [0, 0, 0, 1], equivalent to consider only the slices $ \textbf{S}(:,:,2) $, $ \textbf{S}(:,:,3) $, and $ \textbf{S}(:,:,4)$ respectively; 
	
	\item[Strategy 5:] In this strategy, we consider that  $\boldsymbol{\alpha}^{S5} = [\alpha_1, \alpha_{2}, \alpha_{3}, \alpha_{4}]$ takes on the values: $\alpha_{1}$ = 1 - ($\alpha_{2}$ + $\alpha_{3}$ + $\alpha_{4}$), and $\alpha_{2}, \alpha_{3}, \alpha_{4} \sim U[0.1, 0.2]$, where $U$ is a continuous uniform distribution. We chose to give more importance to the current data and equal weight range for the other features.
	
\end{description}

\begin{table}[htbp]
\centering
\caption{Alternative evaluations in the feature-space.}
\scalebox{0.8}{
	\begin{tabular}{l|rrr|rrr|rrr|rrr}
		\toprule
		Feature & \multicolumn{ 3}{c}{Current data -- $ \textbf{S}(:,:,1)$}
		& \multicolumn{ 3}{c}{Average -- $ \textbf{S}(:,:,2)$}
		& \multicolumn{ 3}{c}{CV -- $ \textbf{S}(:,:,3)$}
		& \multicolumn{ 3}{c}{SC -- $ \textbf{S}(:,:,4)$}\\ \hline
		Criteria & $ c_1 $ &  $ c_2 $ & $ c_3 $ & $ c_1 $ &  $ c_2 $ & $ c_3 $ & $ c_1 $ &  $ c_2 $ & $ c_3 $& $ c_1 $ &  $ c_2 $ & $ c_3 $ \\
		\midrule
		BR    &74,8&11,60&15062  &  70.5  &   9.9 &  12283 
		&0.05    &  0.12 &  0.15
		&  1.9 &  0.70 &  1035
		\\
		
		CN     &76,0&10.30&13347                 &  72.5  &   8.5 &   6004
		&0,03    &  0.15 &  0.69
		&  1.5 &  0.75 &  2336  
		\\
		
		IN     &68.4&8.55&5814                 &  63.4  &   6.9 &   3312
		&0.06    &  0.17 &  0.43
		&  2.1 &  0.68 &   810
		\\
		
		ID    &69,1&10.30&10130              &  66.5  &   8.7 &   6753
		&0.03    &  0.15 &  0.29
		&  1.1 &  0.76 &  1063
		\\
		
		\rowcolor{yellow}
		MY    &74,8&11.35&23712                &  72.9  &  10.0 &  16384
		&0.02    &  0.12 &  0.27
		&  0.8 &  0.67 &  2606
		\\
		
		MX   &77,0&10.80&16249                  &  74.4  &   9.5 &  14137
		&0.03    &  0.10 &  0.11
		&  1.2 &  0.58 &   893  
		\\
		
		PH   &68.3&10.20&8232               &  66.9  &   9.5 &   5805
		&0.01    &  0.05 &  0.28
		&  0.6 &  0.29 &   929 
		\\
		\rowcolor{b}
		RU    &70.3&13.35&22094                 &  67.3 &  12.1 &  17561
		&0.03    &  0.08 &  0.22
		&  0.6 &  0.56 &  1292  
		\\
		
		ZA   &57.9&11.75&12110           &  57.2  &  10.9 &  10691
		&0.06    &  0.08 &  0.09
		& -1.3 &  0.48 &   532  
		\\
		\rowcolor{RawSienna}
		TR   &75.6&11.05&18976                 &  70.6  &   8.8 &  14181
		&0.06 &  0.18 &  0.21
		&  2.3 &  0.93 &  1718
		\\
		\bottomrule
	\end{tabular}}
	\label{tab:atributos} 
\end{table}

The first row of Table~\ref{tab:ranking}, which we called R1, shows the ranking obtained in~\cite{banamar2018extension}, in order to compare with our results. We highlight that this latter study considered the time domain for ranking the countries. The other rows of Table~\ref{tab:ranking} show the ranking achieved by applying our five strategies. To facilitate the analysis, we shall focus on the position of the three countries highlighted in the table: Russia, Malaysia, and Turkey. A first remark on Table~\ref{tab:ranking}  is that the R1 ranking is similar to $\boldsymbol{\alpha}^{S1}$ (current data) and $\boldsymbol{\alpha}^{S2}$ (average). For instance, the ranking in R1, $\boldsymbol{\alpha}^{S1}$, and $\boldsymbol{\alpha}^{S2}$, Russia and Malaysia ranked first and second place. The leading position of these two countries changes when we consider other features of the time-series, as in Strategies 3 and 4.
  
Different rankings were obtained in the results using $\boldsymbol{\alpha}^{S1}$, $\boldsymbol{\alpha}^{S2}$, $\boldsymbol{\alpha}^{S3}$, and $\boldsymbol{\alpha}^{S4}$, as shown in Table~\ref{tab:ranking}. We point out that these four strategies give all the importance to only a specific feature. The $\boldsymbol{\alpha}^{S1}$ (current data) ranking is very close to that of $\boldsymbol{\alpha}^{S2}$ (average). It is possible to see that Russia and Malaysia are in first and second place in both strategies; Turkey, however, changes the position with Mexico. As can be seen in Table~\ref{tab:atributos}, this change of position between Turkey and Mexico is because Turkey presents a better performance in current data, then it is better placed in $\boldsymbol{\alpha}^{S1}$. But in terms of average, Turkey's performance is worse, which leads Turkey to rank worse than Mexico in $\boldsymbol{\alpha}^{S2}$. The rankings in Strategies 3 and 4 are very different than the rankings in Strategies 1 and 2. Indeed, in $\boldsymbol{\alpha}^{S4}$, Russia ranks eighth, rather than first, and Turkey and Malaysia outperform it. This last analysis is clearer to understand from Table~\ref{tab:atributos}, where we observe that in terms of the slope coefficient, Russia's performance is worse than the two countries. 

Given this first analysis, we can show, with a practical example, the discussion presented in Section~\ref{sec:flex}. Here, we note that Russia ranks first in the strategies that consider only the time-space. Instead, when dealing with SC, Russia becomes close to the last place. That is, when we consider the feature-space, a new solution is achieved, and it may be more useful if these elements are relevant for the decision.
		
\begin{table}[htbp]
	\centering
	\caption{Ranking obtained according to the strategy used.}
	\centering
	\scalebox{0.8}{
		\begin{tabular}{lllllllllll}
			\toprule
			{}  &  \nth{1} & \nth{2}  & \nth{3}  &  \nth{4}  & \nth{5}  & \nth{6}  & \nth{7}  & \nth{8}  &  \nth{9}  & \nth{10} \\
			\midrule
			R1 &  \colorbox{b}{RU}  &  \colorbox{yellow}{MY} &  MX &  BR & \colorbox{RawSienna}{TR}  & CN  &  ZA & PH & ID  & IN  \\
			
			$\boldsymbol{\alpha}^{S1}$ & \colorbox{b}{RU}  & \colorbox{yellow}{MY}  & \colorbox{RawSienna}{TR}  & MX  & BR  &  CN &  ZA & ID  &  PH &  IN \\
			
			$\boldsymbol{\alpha}^{S2}$& \colorbox{b}{RU} & \colorbox{yellow}{MY}   & MX  & \colorbox{RawSienna}{TR} & BR  &  ZA &  ID & PH &  CN &  IN  \\		
			
			$\boldsymbol{\alpha}^{S3}$&MX & ZA&BR &\colorbox{b}{RU} & \colorbox{RawSienna}{TR}  &PH   & \colorbox{yellow}{MY} &ID   & IN &  CN  \\
			
			$\boldsymbol{\alpha}^{S4}$& \colorbox{RawSienna}{TR} & CN  & BR  & \colorbox{yellow}{MY} & IN  & ID & MX & \colorbox{b}{RU} & PH   & ZA   \\
			
			$\boldsymbol{\alpha}^{S5}$&\colorbox{yellow}{MY}   & \colorbox{b}{RU}  & \colorbox{RawSienna}{TR}  & BR  & MX  & CN   & ID  &  ZA & PH  & IN   \\
			\bottomrule
		\end{tabular}}
		\label{tab:ranking}  
\end{table}

For the sensitivity analysis, we show in Table~\ref{tab:smaa} the SMAA ranking percentage obtained using $\boldsymbol{\alpha}^{S5}$. As can be seen in Table~\ref{tab:smaa},  in 92\% of the time, Malaysia ranked first.  Russia and Turkey are competing in the second and third place. Russia was ranked 50.19\% of the time in the second place and 49.81\% in the third, while Turkey was 41.88\% in second place and 50.19\% in the third. By analyzing the features in Table~\ref{tab:atributos}, it is possible to see that, in terms of average and current data, Russia performs better than Turkey in criteria $c_2$ and $c_3$. Also,  Russia's coefficient of variation is lower than Turkey's in criteria $c_1$ and $c_2$. Finally, Russia performs worse  in the slope coefficient compared with Turkey for all criteria. Thus, we can infer that the slope coefficient is strongly influencing the dispute between Russia and Turkey for the second and third place. Actually, in $\boldsymbol{\alpha}^{S4}$ (in which all the importance is for  slope coefficient feature), Turkey is in the first position, and Russia is in the eighth. Conversely, in $\boldsymbol{\alpha}^{S1}$, $\boldsymbol{\alpha}^{S2}$, and $\boldsymbol{\alpha}^{S3}$, Russia rank better than Turkey. 

In other pairwise comparisons, like between Mexico and Brazil, the slope coefficient also strongly impacts the dispute to be better placed. As can be seen in Table~\ref{tab:smaa}, Mexico and Brazil are competing for the fourth and fifth places. Brazil was ranked 55.95\% of the time in fourth place and 44.05\% in the fifth, while Mexico was 44.05\% in second place and 55.95\% in the fifth. By investigating the features in Table~\ref{tab:ranking}, Mexico rank better than Brazil in  $\boldsymbol{\alpha}^{S1}$, $\boldsymbol{\alpha}^{S2}$, and $\boldsymbol{\alpha}^{S3}$, but it has a worse $\boldsymbol{\alpha}^{S4}$ performance. Thus, Mexico ranks better than Brazil in three of four features. However, when all features are considered (i.e., in $\boldsymbol{\alpha}^{S5}$), Mexico is not better placed than Brazil; instead they are disputing the fourth and fifth place. 

\begin{table}[htbp]
\caption{Percentage matrix of rankings.}
\centering
\scalebox{0.8}{
	\begin{tabular}{lrrrrrrrrrr}
		\toprule
		{}  &  \nth{1} & \nth{2}  & \nth{3}  &  \nth{4}  & \nth{5}  & \nth{6}  & \nth{7}  & \nth{8}  &  \nth{9}  & \nth{10} \\
		\midrule
		BR & 0   &  0  &   0  &  55.95 & 44.05 &  0  &   0 &    0   &  0  &   0  \\
		
		CN & 0   &  0  &   0  &  0 & 0 &   90.07  &   9.75 &    0.18   &  0  &   0  \\
		
		IN & 0   &  0  &   0  &  0 & 0 &   0  &   0 &   24.22 & 37.09 & 38.69   \\		
		
		ID & 0   &  0  &   0  &  0 & 0 &   0  &   56.09 & 43.91  & 0 & 0  \\
		
		MY&92.07 &  7.93  &   0  &  0 & 0 &   0  &   0 &   0 & 0 & 0  \\
		
		MX& 0   &  0  &   0  &  44.05 &  55.95&   0  &   0 &   0 & 0 & 0  \\
		PH& 0   &  0  &   0  &  0 & 0 &   0  &   0 &   0 & 38.75 & 61.25  \\
		RU& 0   &     50.19 &  49.81 & 0 & 0 &   0  &   0 &   0 & 0 & 0  \\
		ZA& 0   &  0  &   0  &  0 & 0 &   9.93&  34.16&   31.69 &  24.16&    0.06  \\
		TR& 7.93 & 41.88 & 50.19   &  0 & 0 &   0  &   0 &   0 & 0 & 0  \\
		\bottomrule
	\end{tabular}}
	\label{tab:smaa}  
\end{table}

It is worth noticing that the good performance of some countries in terms of current data, average, and coefficient of variation, was compensated by the worst performance in the slope coefficient. In this sense, the ranking proved to be very sensitive to the value of the $\boldsymbol{\alpha}$. Therefore, it is crucial to analyze how important is the tendency for decision-making.   

\section{Conclusion}\label{sec:conclusoes}

This paper structure the MCDA data in a tensorial approach to fully exploit relevant features for decision-making. We proposed an extension of the TOPSIS method for aggregate the tensor to rank the alternatives. We also provided a sensitivity analysis using the SMAA method to verify the impact of the feature's weights in the ranking. For the computational tests, we applied the approach in a real-word time-series of ten countries to ranking them according to three criteria. 

The main conclusion of the analysis is that considering some features from the time-series may lead to a different perspective in the decision-making compared to only aggregating the time-series or the current data. This approach allows us to find new solutions (rankings) that better describe the decision problem. 

From the experiments, we observe changes in the ranking when considering new elements beyond the current data. It was possible to see that the tendency of the time-series strongly impacts the final ranking. For example, it is interesting if the consequences of the decision are in the medium or long term. Thus, we conclude that the feature-space approach can be meaningful for the decision-maker.

\bibliographystyle{elsarticle-num}

\bibliography{bibliografia_doutorado}

\end{document}